%% file: ssrlda-emnlp14-1014.tex
\documentclass[11pt]{article}
\usepackage{acl2014}
\usepackage{times}
\usepackage{multicol} 
\usepackage{url}
\usepackage{latexsym}
\usepackage{amsmath,amsfonts,amssymb}
\usepackage{multirow}
\usepackage{graphicx,graphics}
\usepackage{color}

\newcommand{\Dir}{\mathrm{Dirichlet}}

\newcommand{\ADJ}{{\sc adj}~}
\newcommand{\ADV}{{\sc adv}~}
\newcommand{\ENTL}{$\text{{\sc ent}}^{left}$~}
\newcommand{\ENTR}{$\text{{\sc ent}}^{right}$~}
\newcommand{\NN}{{\sc nn}~}
\newcommand{\OTH}{{\sc oth}~}
\newcommand{\PP}{{\sc pp}~}
\newcommand{\VB}{{\sc vb}~}
\newcommand{\POSSEQ}{{\sc pos-seq}~}
\newcommand{\TYPE}{{\sc ent-type}~}

\newcommand{\PER}{{\sc per}~}
\newcommand{\ORG}{{\sc org}~}
\newcommand{\LOC}{{\sc loc}~}
\newcommand{\MISC}{{\sc misc}~}

\newcommand{\AQUAINT}{{\sc Aquaint}2~}

\newcommand{\Ehat}{\hat{\mathbb{E}}}
\newcommand{\E}{\mathbb{E}}
\newcommand{\mathbackslash}{\text{\textbackslash}}

\makeatletter
\let\memoir@makechapterhead\@makechapterhead
\def\@makechapterhead#1{\memoir@makechapterhead{#1}\dochapterprolog}
\makeatother

\def\chapterprolog#1{\gdef\dochapterprolog{%
  \if\relax\detokenize{#1}\relax\else
    \noindent\begin{minipage}{\textwidth}#1\end{minipage}\vspace{3\baselineskip}
  \fi\chapterprolog{}}}

\title{Online Inference for Relation Extraction with a Reduced Feature Set}

\author{Maxim Rabinovich \\
Computer Science Division\\
University of California, Berkeley \\
{\tt rabinovich@eecs.berkeley.edu} \\\And
C\'{e}dric Archambeau\thanks{${}^{\ast}$ Work undertaken while the second author was at Xerox Research Centre Europe, supervising the first author's research internship.} \\
Amazon\\
Berlin, Germany \\
{\tt cedrica@amazon.de} \\}

\date{}

\begin{document}
\maketitle
\begin{abstract}
{Access to web-scale corpora is gradually bringing robust automatic knowledge base creation and
extension within reach. To exploit these large unannotated---and extremely difficult to annotate---corpora,
unsupervised machine learning methods are required. Probabilistic models of text have recently found 
some success as such a tool, but scalability remains an obstacle in their application, with standard approaches
relying on sampling schemes that are known to be difficult to scale. In this report, we therefore present an empirical assessment of the sublinear time sparse stochastic variational inference (SSVI) scheme applied to RelLDA. We demonstrate that online inference leads to relatively strong qualitative results but also identify some of its pathologies---and those of the model---which will need to be overcome if SSVI is to be used for large-scale relation extraction.
}
\end{abstract}

\section{Introduction}

Access to web-scale corpora is gradually bringing automatic knowledge base creation and
extension within reach \cite{mausam12ollie}. Human curated resources, such as Freebase \cite{bollacker08free}, are invaluable for relation extraction, but they are inherently incomplete. The total number of relations that might be encountered is unbounded and the number actually encountered in a corpus grows with its size. Hence the need for unsupervised methods and the recent small-scale success on this problem with probabilistic models. Unfortunately, prohibitive memory usage and training time makes their large-scale application all but impossible, and the incremental training algorithms used to train topic
models like latent Dirichlet allocation (LDA) at scale \cite{hoffman10stochlda,mimno12sparse} have not yet been applied to relation extraction.


In this paper, we show that sparse stochastic variational inference (SSVI) \cite{mimno12sparse}
can be applied to the RelLDA model for unsupervised relation extraction introduced by \cite{yao11rlda,yao12sense}.  SSVI is attractive for two reasons. First, it processes corpora incrementally, speeding convergence and supporting streaming. Second, it improves on plain stochastic variational inference by using
sparse updates able to deal with a large number of topics. We find that our algorithm is able to obtain strong qualitative results in a fraction of the time that is needed
to run the Gibbs sampler for RelLDA and with a reduced memory footprint. We also include discussion of some pitfalls in unsupervised relation extraction with LDA-style models and
how they might be overcome, and we show that dependency parse features are not needed for this task, a major departure from prior work in this area. 



\section{Model Specification}


We use a modified form of RelLDA \cite{yao11rlda}, eliminating the reliance on a dependency parsed corpus. Relations are grouped into clusters. Each document is assumed to behave as a mixture of these relation clusters, with each sentence in the document exhibiting exactly one of them. Multiple feature sets are permitted, which we exploit below to use separate vocabularies for entity features, linking word features, and syntactic features. Throughout this paper, we adopt the convention that $R$ refers to the number of relation clusters, $F$ to the number of feature types, $W_{f}$ to the vocabulary size for feature type $f~ (1 \leq f \leq F)$, $N_{d}$ to the
number of sentences in a document, and $N_{dif}$ to the number of features of type $f$ exhibited by sentence $i$ in document $d$.

In this notation, the relation clusters are defined as a set of $F$ discrete distributions over the feature vocabularies:
\begin{description}

\item[] For $r = 1, \dots, R$ and $f = 1, \dots, F$: \\ Draw $\beta_{rf} \sim \Dir(\eta_{f})$,

\end{description}
where $\eta_{f} > 0$ is a scalar.\footnote{Note that this means we are using a symmetric Dirichlet, viz. $p(\beta~|~\eta) \propto \prod_{v}{\beta_{v}^{\eta - 1}}$.} The generative process for relations takes the following form:

%
%
%
%
%
%
%
%
%
%

\begin{description}

\item[] For $d = 1, \dots, D$:

    \begin{enumerate}

    \item
    Draw $\theta_{d} \sim \Dir(\alpha)$.

    \item
    For $i = 1, \dots, N_{d}$:
        \begin{enumerate}

        \item 
        Draw $z_{di} \sim \theta_d$.

        \item
        For $f = 1, \dots, F$ and $j = 1, \dots, N_{dif}$: Draw $w_{dij} \sim \beta_{z_{di}f}$.

        \end{enumerate}

    \end{enumerate}

\end{description}
where $\alpha > 0$ is again a scalar, and $\theta_d$ defines a discrete distribution over the relation clusters associated to document $d$. The relation in each sentence is drawn from $\theta_d$ and the associated features from $\beta_{rf}$.


\subsection{Extracting entity pairs}


We assume access to a part-of-speech (POS) tagger and a named entity recognizer (NER). Our ultimate goal is to extract relations between named entities and therefore necessarily limit attention to sentences with at least two entity mentions. Sentences with more than two mentions pose a problem due to \textit{a priori} ambiguity in the pairs being related, so we simply assume the salient entity pair is the one that is closest together in the sentence---a simple heuristic that allows us to avoid modeling sentence segmentation. We use the Stanford CoreNLP library for both POS tagging and NER \cite{finkel05ner}.

\subsection{Feature sets}

Our experiments all draw on feature types built according to a small set of templates and always reflecting only the sequence of words \textit{between} two 
selected entity mentions in the sentence:

\begin{description}

\item[Entity surface strings.] Each sentence contains two distinguished entity mentions. The left (first) and right (second) strings are treated as features of distinct types to capture asymmetry. The vocabularies for those two types are, however, the same. We refer to
the resultant features as \ENTL and \ENTR.

\item[Entity types.] The Stanford NER outputs entity types in \{ \PER, \ORG, \LOC, \MISC \}, referring to the {\bf per}son, {\bf org}anization, {\bf loc}ation, and {\bf misc}ellaneous, respectively. We use the pair $(t_1,~ t_2)$ of entity types for the two distinguished entities as a feature. This feature type is 
referred to as \TYPE.

\item[Phrases between the entities.] The word sequence between the entities is partitioned into coarse-grained part-of-speech categories: \ADJ  (JJ, JJR, JJS), \ADV  (RB, RBR, RBS), \NN  (NN, NNS, NNP, NNPS, PRP, WP), \PP  (IN, TO), \VB  (VB, VBD, VBG, VBN, VBP, VBZ), and \OTH  (everything else). 
We refer to the resultant
six feature sets as \ADJ, \ADV, \NN, \OTH, \PP, and \VB. 

\item[POS tag sequences.] We include a feature corresponding to the entire sequence of Penn Treebank POS tags between the two entities. We refer to this feature type as \POSSEQ. 

\end{description}

\section{Sparse Stochastic Variational Inference}


To make inference scalable to very large corpora, we use the sparse stochastic variational inference (SSVI) originally developed for LDA \cite{mimno12sparse}. The true posterior over $\beta_{1:R,1:F}$ is approximated by a product of independent Dirichlets, viz.
$$ q(\beta_{1:R,1:F}) = \prod_{r = 1}^{R}\prod_{f = 1}^{F} q(\beta_{rf}), $$
where $ q(\beta_{rf}) = \Dir(\lambda_{rf})$ and $\lambda_{rf} \in \mathbb{R}_{+}^{W_{f}}$ are variational parameters. Classical variational Bayes would also approximate the posterior over $\theta_{d}$ and $z_{di}$ by Dirichlet and multinomial distributions, respectively, leading to $\Omega(DR)$ memory usage and $\Omega(R)$ time for local updates. SSVI reduces both requirements to $O(1)$ by eliminating the local variational distribution. Instead, it integrates out $\theta_d$ and uses samples from the an optimized variational distribution $q^{\ast}(z_d)$ to estimate the expectations required in the updates. Here the
optimality criterion for $q^{\ast}$ is simply that its Kullback-Leibler divergence from the true $z_d$ posterior is as small as possible within the constraints
imposed by its factored form \cite{bishop06prml}. 



Furthermore, the entire corpus need not be considered
during each step; rather, a random minibatch $B = \lbrace d_{1}, \dots, d_{S} \rbrace$ of documents is considered and sampling is carried out only for those documents. Each iteration thus only needs to update the parameters associated with relations $r$, features types $f$, and features values $v$ encountered in 
$B$. This leads to the following variational updates:
$$ \lambda_{rfv}^{(t + 1)} = (1 - \rho^{(t)})\lambda_{rfv}^{(t)} + \rho^{(t)}\cdot \frac{D}{S}\sum_{d \in B}{\Ehat\left[N_{drfv}\right]}, $$
where $\rho^{(t)}$ is the learning rate, $N_{drfv}$ is the number of times feature value $v$ of type $f$ is assigned to relation $r$ in document $d$ and $\Ehat$ denotes a Monte Carlo estimate of an expectation. Using a trick we explain in the supplement, we can ensure that each iteration only updates parameters $\lambda_{rfv}$ for relations $r$, feature types $f$, and feature values $v$ that occur in that iteration's minibatch (the origin of the \textit{sparse}
moniker). The supplement likewise explains our natural gradient hyperparameter optimization scheme for $\eta_{f}$ and $\alpha$. 


\section{Empirical Evaluation}

\subsection{Datasets}

We use the \AQUAINT2 corpus, consisting of articles from several newspapers including the New York Times \cite{aquaint2}. After eliminating sentences with fewer than two entities, we were left with 578790 documents (1492599 sentences), of which 462755 (1193275 sentences) were used in training and the remainder used for 
evaluation. The sizes of the feature sets for this data were: 8996 (\ADJ), 7334 (\ADV), 233725 (\ENTL), 233725 (\ENTR), 39895 (\NN), 52998 (\OTH), 16564 (\PP), 28826 (\VB), 89022 (\POSSEQ), and 16 (\TYPE). We consider two subset of the features in our experiments: 

\begin{enumerate}

\item
The full feature set: \ADJ, \ADV, \ENTL, \ENTR, \OTH, \PP, \VB, \POSSEQ, and \TYPE.

\item
All features excluding the entity features: \ADJ, \ADV, \OTH, \PP, \VB, \POSSEQ, and \TYPE. 

\end{enumerate}


\subsection{Model selection}

\begin{figure*}[t]
\centering
\includegraphics[width=3.5in,height=2.5in]{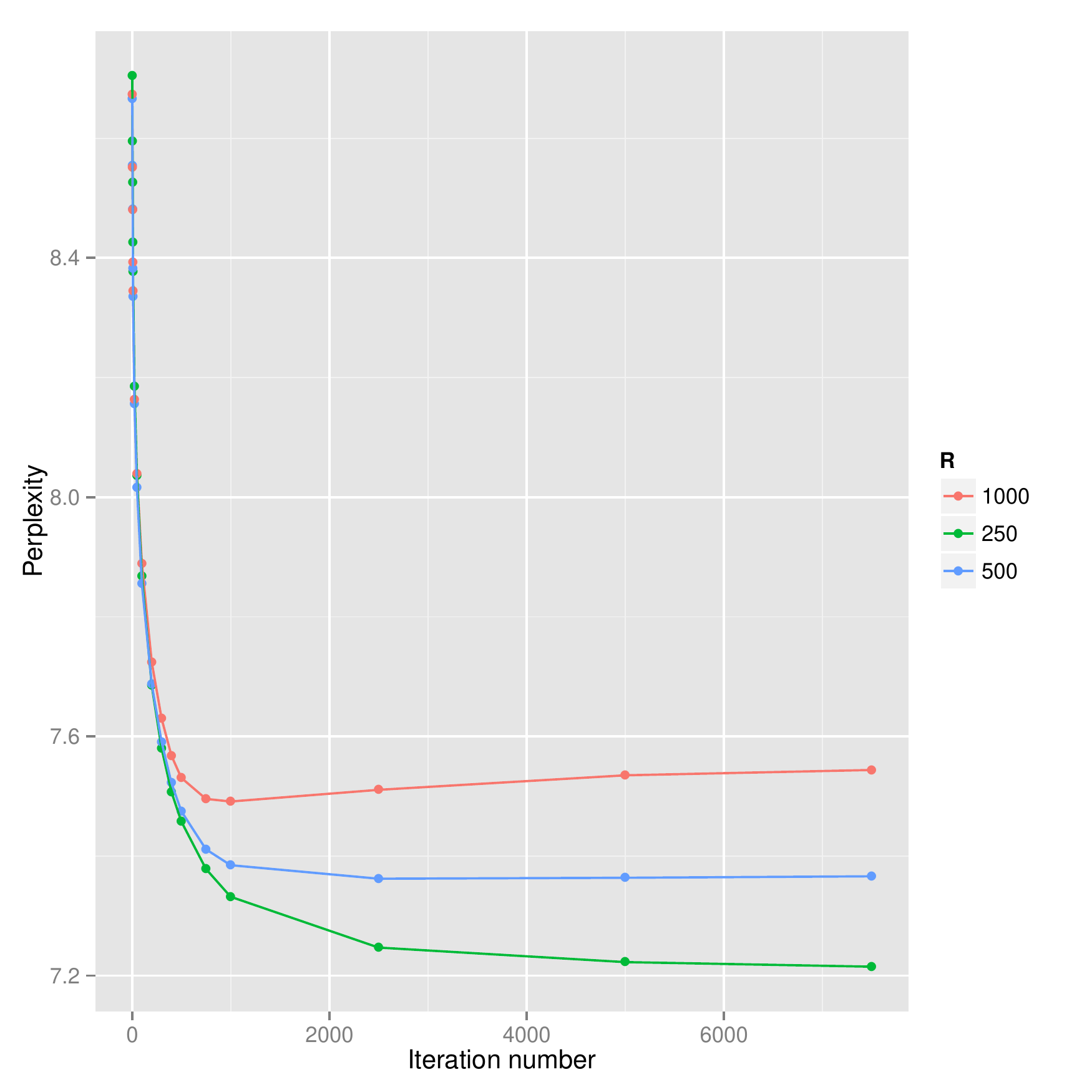}\includegraphics[width=3.5in,height=2.5in]{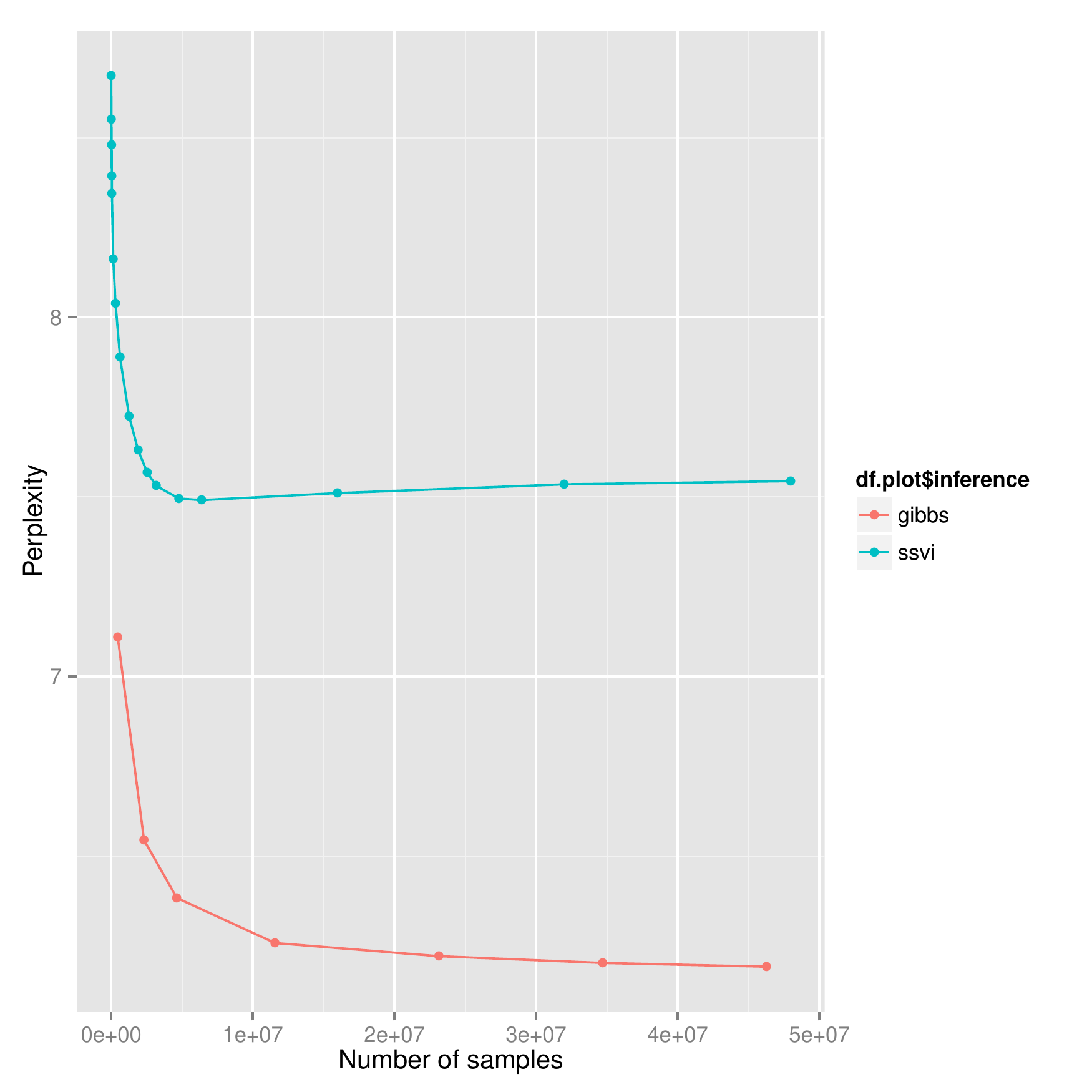}
\caption{\emph{(Left)} Perplexity on an evaluation corpus for SSVI as a function of iteration ($a = 0.01,~ b = 1.0$). \emph{(Right)} A comparison of evaluation perplexity for SSVI and Gibbs sampling with $R = 1000$.}

\label{fig:iter-select}
\end{figure*}


The hyperparameters are optimized as part of the algorithm. SSVI includes a learning rate $\rho^{(t)}$ generally set to
$$ \rho^{(t)} = \frac{a}{\left(b + t\right)^{c}} ,$$
where $a, b > 0$ and $\frac{1}{2} < c \leq 1$. This choice of schedules allows convergence of the algorithm to a local optimum of the objective to be guaranteed \cite{hoffman13stochvar}. 
In practice, setting $c$ at or close to $\frac{1}{2}$ give good results. 

We fit the model with several values of $R, a$, and $b$ and score each based on its perplexity and variational objective values on an evaluation corpus. We carried out a grid search for values with 
$R \in \lbrace 250, 500, 1000 \rbrace$, $a \in \lbrace 0.1, 0.01, 0.001 \rbrace$, and $b \in \lbrace 1.0, 10.0 \rbrace$. We find that the choice of these parameters has a noticeable but not substantial effect on the metrics. 
Nonetheless, we limited our qualitative evaluation to the best learning rates in terms of the variational objective ($-9.02\times 10^6$), that is, $a = 0.01, b = 10.0$ and $K = 500$. The number of iterations $T$ of SSVI, on the other hand, had a substantial effect. Figure \ref{fig:iter-select} illustrates this with varying values of $R$ and $a = 0.1, ~b = 1.0$.




\subsection{Discovered relations}

%


\begin{table*}[ht]
\hspace{-0.75in}
\scalebox{0.65}{
\begin{tabular}{c}
\texttt{leader-at} relation \\ \hline
European / Peter Mandelson / trade commissioner / NN NN OT / MISC-PER \\
UN / Joao Bernardo Honwana / special envoy / NN NN / ORG-PER \\
UN / Pierre Goldschmidt / director general / 's deputy / ORG-PER \\
Zimbabwe / Morgan Tsvangirai / opposition leader / NN NN / LOC-ORG \\
Spanish / Jose Antonio Alonso / counterpart , / NN OT / MISC-PER \\
European Union / Pascal Lamy / trade commissioner / NN NN / ORG-PER \\
ASIO / Dennis Richardson / director general / OT NN JJ / ORG-PER \\
WTO / EU / trade commissioner / NN NN OT / MISC-PER \\
UN / Jacques Klein / special envoy / NN NN / ORG-PER \\
pro-Russian / Viktor Yanukovich / opposition leader / NN NN / MISC-PER
\end{tabular}
\begin{tabular}{c} 
\texttt{trader-at}/\texttt{market-strategist-at} relation \\ \hline
Roma / Livorno / bottom club / 2-0 away to / VBD CD RB TO NN NN / ORG-LOC \\
Art Hogan / Jefferies and Co. / market strategist at / , chief / OT JJ NN NN IN / PER-ORG \\
Kenneth Tower / CyberTrader / market strategist at / , chief / OT JJ NN NN IN / PER-ORG \\
Chinese / Ssangyong / bidder for / firm , / as the / NN OT VBD IN DT JJ NN IN / MISC-ORG \\
Michael Sheldon / Spencer Clark LLC / market strategist at / , chief / OT JJ NN NN IN / PER-ORG \\
Oracle Corp. / PeopleSoft / business software maker / bid for / ORG-ORG \\
US / Asian / market strategist at / , chief / OT JJ NN NN IN / PER-ORG \\
SAIC / Birmingham-based / automaker , / fortunes of / one billion / that could potentially / 1.85 billion / ORG-MISC \\
Barry Ritholtz / Maxim Group / market strategist at / , chief / OT JJ NN NN IN / PER-ORG \\
Al Goldman / AG Edwards / market strategist at / , chief / OT JJ NN NN IN / PER-ORG \\
\end{tabular}}
\caption{Sentences in the corpus most strongly associated with one of the relations, as determined by sampling relation assignments. \emph{(Top)} This particular relation appears to identify the concept of ``occupies leadership position at,'' while the second relation \emph{(bottom)} appears to identify the concept of ``trader at'' or ``trading strategist at.''  Parameters were set to $R = 500$, $a = 0.009$, and $b = 10.0$ for both.}
\label{tab:rel-coherent}
\end{table*}

Evaluating the quality of the relations discovered by our algorithm is challenging in the absence of ground truth, especially due to the inherent noisiness of relation clusters discovered by any unsupervised learning algorithm---and by stochastic gradient methods in particular. 
Ordinarily, the output of LDA-type models is shown as per-topic rankings of the vocabulary. In our setting, this makes little sense due to the multi-view setup and the fact that, e.g., the most likely entities under a relation need not correspond to the most likely noun phrases. We thus represent relation clusters as lists of sentences most strongly associated with them. The strength of association was determined by taking $50$ posterior samples of the relation assignment for each sentence and computing the proportion of samples assigned to each relation. 

As Table \ref{tab:rel-coherent} shows, the clusters are reasonably coherent but quite noisy. The first corresponds to a general constellation of relations between people and organizations that could reasonably be summarized as ``occupies leadership
position at," though in reality,  the generalization made by the inference procedure is somewhat narrower than that, with a bias toward political leaders. The second is much more restricted and basically corresponds to the concept of being a ``market strategist at.'' Even so, the model picks up on the fact that ``bidder for'' is 
a closely related concept and expresses a similar relationship between the person and organization in question. 

\subsection{Clustering pathologies}

\begin{figure*}[t]
\begin{tabular}{c|c} 
Entity-based relation & Topic-like relation \\ \hline
~~ & ~~ \\
French Riviera / Cannes / resort of & policemen and / city of /near the / were killed \\
Northern Gaza / Israeli / withdrawal of the / and the & blows himself / suicide bomber / up during / when a  \\
Gaza / Israeli  / withdrawal of the / and the & incursion into / the northern  \\
France / China / deficit with & rebel stronghold of / roadside bomb / were wounded \\ 
~~ & ~~\\ 
\end{tabular}
\caption{\emph{(Left)} A relation based on sets of related entities. \emph{(Right)} A topic-like relation. Both exclude \POSSEQ and \TYPE features for brevity.}
\label{fig:bad-rel-ex}
\end{figure*}

The results we show correspond to a feature set excluding \ENTL and \ENTR, as we found certain pathologies in the output with the full feature set, 
notably a tendency for some relation clusters to form around sets of entities rather than the relations between them. Figure \ref{fig:bad-rel-ex} illustrates this
effect.

Removing entity features resolves this first issue. Overcoarsening of relation clusters is a more persistent problem. Some relations look more like broad topics than
focused relations. The likely cause of this is allocation of topic words that co-occur with relation words to that relation due to the
absence of a special set of shared topic distributions that could catch the intruding words. Figure \ref{fig:bad-rel-ex} illustrates this problem within one relation.

The incorporation of syntactic features is another source of over-coarsening, with some relation clusters forming around common syntactic patterns. 
The best illustration of this are the POS tag sequences ``NNS IN'' and ``NN IN", which served as the basis for clustering of unrelated concepts like 
``headquarters in,'' ``crisis in,'' and ``meeting in.''

At their core, the pathologies we uncover all appear to flow from problems in the model rather than the inference scheme---notably the requirement that each word be explained by a relation. The
absence of any broader shared topic distributions that can be used to explain away non-relation-specific words causes some relations to behave very much
like topics and all relations to catch many co-occurrent words that are not essentially part of their semantic content. Likewise, although the addition of syntactic 
features allows abstraction away from specific word patterns to more generally applicable syntactic ones, it also leads to problems
if the model does not account for syntactic overlap of semantically distinct relations. Both of these issues could be addressed 
by adding additional hierarchy to the model. For instance, a set of global topic distributions could be added to resolve the first problem, while relations could
be grouped into higher-level clusters governing syntactic properties to resolve the other. We believe such modifications are likely to lead to much more robust
models of relations in text without significantly complicating inference. 



\subsection{Comparison to Gibbs sampling}

Since we use a minibatch size of $S = 256$ and $S' = 25$ samples to form our estimate of the natural gradient, each iteration of SSVI corresponds to 
$6400$ document steps in the Gibbs chain. As a result, one full Gibbs sweep through the corpus is equivalent to about $75$ SSVI iterations in terms of
numbers of samples taken.\footnote{A more exact number is $\frac{462755}{6400} \approx 72.3$.} We use this as the basis of the plot in 
Figure \ref{fig:iter-select}. 

Surprisingly, Gibbs sampling appears to achieve better held-out perplexity at each given level of computation. This is contrary to the expected behavior of SSVI \cite{hoffman10stochlda,mimno12sparse} and does not have a clear explanation. The most likely causes are, first, the learning parameters, as stochastic gradient methods are known to be extremely sensitive to the choice of learning rate \cite{ranganath13svi} and, second, the inherent noisiness 
of stochastic gradient methods, which work best on large, highly redundant corpora. Although we lightly optimized the learning parameters, it is possible that more extensive experiments would discover a drastically better setting of those parameters; alternatively, adaptive rate methods may be needed. If,
on the other hand, is simply the noisiness of the stochastic gradients, then variance reduction techniques may yield better results \cite{paisley12control}.

It is also important to account for the computational aspect of the performance metric. Often, one can drastically reduce the size of the minibatches in SSVI (e.g. to $64$ documents), which would lead to multiplicative speedups (e.g. $4$x if $S = 64$); likewise, the number of Gibbs sweeps used for 
estimation on the minibatch could be reduced, as could the burnin for those sweeps. Such fine-tuning is beyond the scope of this work but, based on our 
results, would be a crucial component in practical systems seeking to reap the benefits of SSVI with models like RelLDA. 

Finally, it may simply be that for a complex model like RelLDA, the data set must be made far larger before sufficient redundancy appears, in which case we would expect to see gains in the relative performance of SSVI and Gibbs sampling in the regime of larger information extraction datasets, which often
contain hundreds of millions of documents. This last point also illustrates how SSVI might be advantageous even if less statistically efficient: unlike Gibbs
sampling, whose memory usage grows with the size of the corpus, SSVI can operate with a fixed amount of memory---just enough to store the minibatch
data structures. 

\section{Conclusion}

We have shown that SSVI is a promising technique for relation extraction at scale. Apart from some pathologies due to the modeling assumptions, it discovers coherent relational
clusters while requiring less memory and time than sampling methods. Moreover, the issues we uncover point to problems with the model that 
suggest how more effective probabilistic models of relations in text might be designed and used.


\section*{Acknowledgements}

The author would like to thank Guillaume Bouchard for helpful discussions about improvements to the core model.

\input{supp-include}

\bibliography{ssrlda-emnlp14-1014}
\bibliographystyle{acl}



\end{document}

%% file: supp-include.tex







\appendix
\section*{Appendices}
\noindent In the following appendices, we explain the mathematics of our inference algorithm in detail.

\section{The core algoritihm}

An alternative to MAP inference via Gibbs sampling is variational inference. Ordinarily, this would be done by specifying a variational distribution over the $\beta$, $z$, and 
$\theta$ variables. In our setup, because each observation consists of multiple words, the standard way of doing this fails, however. Fortunately, we can use a recent stochastic 
approach that still works and that scales much better than batch variational Bayes. This approach is based on the strategy for LDA set out in 
\cite{mimno12sparse}. 

To do this, we posit
$$ q(\beta_{rf}) = \mathrm{Dir}(\lambda_{rf}),~~ \lambda_{rf} \in \mathbb{R}_{+}^{V_f} $$
and let $q(z_d)$ be an arbitrary distribution, which will be chosen to be the optimal one per the analytical (but uncomputable) variational Bayes update formula. The mixing
distributions $\theta$ are marginalized out as in collapsed Gibbs sampling. Since our goal is to optimize $\lambda$, we write the ELBO up to a constant independent of $\lambda$:
\begin{align*}
\mathcal{L} & = \sum_{d}\sum_{r,f}\Bigg[ \sum_{v}\left(\E\left[N_{drfv}\right] + \frac{\eta_f - \lambda_{rf}}{D}\right) \\       
                  &~~~~~~~~~~~~~~~~~~~~~~~~~~~~~~ \times \E_{q}\left[\log\beta_{rfv}\right]  \\ 
                  &~~~~~~~~~~ + \frac{1}{D}\left(\sum_{v}{\log\Gamma(\lambda_{rfv})} - \log\Gamma(\Lambda_{rf})\right)\Bigg],
\end{align*}
where $\Lambda_{rf} = \sum_{v}{\lambda_{rfv}}$. We know $\E_{q}[\log{\beta_{rfv}}] = \Psi(\lambda_{rfv}) - \Psi(\Lambda_{rf})$, and we use sampling over $z_d$ to approximate
$\E_{q}[N_{drfv}]$. Specifically, basic theory tells us that the optimal choice of variational distribution over $z_d$, holding those over all other latent variables fixed, is
\begin{align*}
q^{\ast}(z_d) & \propto \exp\left(\E_{q^{\mathbackslash z_d}}[\log p(z_{1:D},~ v_{1:D},~ \beta_{1:R,~1:F}]\right) \\
              & \propto \exp\left(\E_{q^{\mathbackslash z_d}}\left[\log p(v_{d}~ |~z_d,~ \beta) + \log p(z_{d}~|~\alpha)\right]\right) \\
              & \propto p(z_d~|~\alpha) \prod_{r,~ f}{\prod_{v \colon N_{drfv} > 0}{\exp\left(N_{drfv}\E_{q}\left[\log{\beta_{rfv}}\right]\right)}} \\
              & = \left(\frac{\Gamma(R\alpha)}{\Gamma(O_d + R\alpha)} \cdot \prod_{r}{\frac{\Gamma(O_{dr} + \alpha)}{\Gamma(\alpha)}}\right) \\
              &~~~~~~~~~~ \times \prod_{r,~ f}{\prod_{v \colon N_{drfv} > 0}{\exp\left(N_{drfv}\left[\Psi(\lambda_{rfv}) - \Psi(\Lambda_{rf})\right]\right)}} . 
\end{align*}
We thus find
\begin{align}\label{eq:sparse-var-gibbs-dist}
q^{\ast} & (z_{do} = r~|~z_{d}^{\mathbackslash do}) \propto \\
                        & (O_{dr} + \alpha)  \nonumber \\ 
                        &~~~~~ \times \prod_{f}{\prod_{v \colon N_{dofv} > 0}{\exp\left(N_{dofv}\left[\Psi(\lambda_{rfv}) - \Psi(\Lambda_{rf})\right]\right)}} , \nonumber
\end{align}
which means we can approximately sample from $q^{\ast}$ using Gibbs sampling to obtain an approximation to $\E_{q}[N_{drfv}]$.

Why is this helpful? As shown in \cite{hoffman13stochvar}, the natural gradient of $\mathcal{L}$ in the $rfv$ dimension is given by
$$ \E_{q(z_{1:D})}\left[\sum_{d}{N_{drfv}}\right] + \eta - \lambda_{rfv} . $$
Split up over documents, this gives a per-document contribution of
$$ \E_{q(z_d)}\left[N_{drfv}\right] + \frac{1}{D}\left(\eta - \lambda_{rfv}\right) . $$
This means that if we sample a batch of documents $d_{1}, \dots, d_{S}$ and approximate $\E_{q(z_d)}\left[N_{drfv}\right]$ using $S'$ rounds of Gibbs sampling, we will end up with
an unbiased estimate of the natural gradient that we can use for stochastic gradient ascent on $\mathcal{L}$. With a little bit more work, we can make all necessary updates 
sparse to ensure efficiency. 

Concretely, each iteration of the algorithm does the following. 

\begin{enumerate}

\item
Sample a minibatch $\mathcal{M} = \lbrace d_1, \dots, d_S \rbrace $ of $S$ documents (without replacement). 

\item
Run $B$ burn-in rounds of Gibbs sampling on $z_{d}$ for $d \in \mathcal{M}$ using \eqref{eq:sparse-var-gibbs-dist}. Then run $S'$ more sweeps, saving the value of $\sum_{d \in \mathcal{M}}{N_{drfv}}$ after each one. Estimate $\sum_{d \in \mathcal{M}}\E_{q}[N_{drfv}]$ by $\hat{N}^{\mathcal{M}}_{drfv} \colon = \frac{1}{S'}\sum_{s' = 1}^{S'}{N_{drfv}^{(s')}}$.

\item
Estimate the $rfv$ component of the overall natural gradient by
$$ \hat{g}_{rfv} \colon = \frac{D}{S} \cdot \hat{N}^{\mathcal{M}}_{rfv} + \eta_{f} - \lambda_{rfv} . $$

\item
Update
$$ \lambda_{rfv} \leftarrow \lambda_{rfv} + \rho\hat{g}_{rfv}, $$
where $\rho = \rho_t$ is the current learning rate.

\end{enumerate}

\noindent Note that if we write $\hat{N}_{drfv} = \frac{D}{S} \cdot \hat{N}^{\mathcal{M}}_{rfv}$ and let $\tilde{N}_{rfv} = \lambda_{rfv} - \eta_{f}$ (this is the pseudocount part of the
variational parameter), we have $\lambda_{rfv} = \tilde{N}_{rfv} + \eta_f$ and hence an update of the form
$$ \tilde{N}_{rfv} \leftarrow (1 - \rho)\tilde{N}_{rfv} + \rho\hat{N}_{rfv} . $$
 
Note further that if we let $\pi_t = \prod_{\tau = 0}^{t}{(1 - \rho_{\tau})}$, we can write this update as 
$$ \frac{\tilde{N}_{rfv}^{(t)}}{\pi_t} = \frac{\tilde{N}_{rfv}^{(t - 1)}}{\pi_{t - 1}} + \frac{\rho\hat{N}^{(t)}_{rfv}}{\pi_{t}} . $$
Thus, if we track $\frac{\tilde{N}^{(t)}_{rfv}}{\pi_t}$ rather than the raw pseudocount, we get sparse updates. This is what the code actually does.

\section{Adding hyperparameter optimization}

In its current form, the variational inference algorithm requires the Dirichlet hyperparameters $\eta_{f}$ to the global relation distributions $\beta_{rf}$ 
and $\alpha$ to the local mixing distributions $\theta_{d}$ to be set manually. To remove this limitation, we extend the natural gradient descent scheme
to the hyperparameters. 


To begin, note that the part of the variational objective that depends on $\eta_f$ is given by 
\begin{align*}
\mathcal{L}(\eta_f) & = \eta_f \cdot \sum_{r}{\sum_{v}{\left[\Psi(\lambda_{rfv}) - \Psi(\Lambda_{rf})\right]}} \\ 
                             &~~~~~~~~~~ - R \cdot \left[V_{f}\cdot\log\Gamma(\eta_f) - \log\Gamma(V_f \eta_f)\right] , 
\end{align*}
whence
\begin{align*}
\frac{\partial\mathcal{L}}{\partial \eta_f} & = \sum_{r}\Bigg[\sum_{v}{\left[\Psi(\lambda_{rfv}) - \Psi(\eta_f)\right]} \\ 
                                                                &~~~~~~~~~~ - V_{f}\cdot \left[\Psi(\Lambda_{rf}) - \Psi(V_f \eta_f)\right]\Bigg] . 
\end{align*}

However, we would like to use natural gradient updates, which have the form
$$ \eta_{f}^{(t + 1)} = \eta_{f}^{(t)} + \rho_{t}\left[G_{\eta, f}^{(t)}\right]^{-1}\nabla_{\eta_f}{\mathcal{L}} , $$
where $G_{\eta, f}^{(t)} = \E\left[\left(\frac{\partial \log{p(\beta_{f}~|~\eta_f)}}{\partial \eta_f}\right)^2 ~ |~ \eta_f\right](\eta_{f}^{(t)})$ is the Fisher information matrix for the parameter $\eta_f$ evaluated at the value $\eta_{f}^{(t)}$. Since
\begin{align*}
\log{p(\beta_f ~ |~ \eta_f)} & = (\eta_f - 1) \cdot \sum_{r, v}{\log \beta_{rfv}} \\
                                          &~~~~~~~~~~ - R \left(V_{f}\log{\Gamma(\eta_f)} - \log{\Gamma(V_f \eta_f)}\right) . 
\end{align*}
This is easy to compute. 

Indeed, if we write $\log{p(\beta_f~|~ \eta_f)} = t(\beta_f)\cdot\eta_f - t(\beta_f) - a(\eta_f)$ with $t(\beta_f) = \sum_{r}{\sum_{v}{\log{\beta_{rfv}}}}$, 
we need only compute $\E\left[\left(t(\beta_f) - a'(\eta_f)\right)^2\right]$, which, by the usual exponential family identities, is given by
$$ \E\left[t(\beta_f)^2\right] - \E\left[t(\beta_f)\right]^2 = a''(\eta_f) . $$
Fortunately, we know
$$ a'(\eta_f) = RV_{f}\cdot \left[\Psi(\eta_f) - \Psi(V_{f}\eta_f)\right] , $$
so we can calculate
$$G_{\eta,f}(\eta_f) = a''(\eta_f) = RV_{f}\cdot\left[\psi_1(\eta_f) - V_{f}\psi_1(V_f\eta_f)\right] , $$ 
where $\psi_1 = \Psi'$ is the first polygamma function (the trigamma function). Note that, analogously,
$$ G_{\alpha}(\alpha) = DR\cdot\left[\psi_1(\alpha) - R\psi_1(R\alpha)\right] . $$

The (unnatural) gradient for $\alpha$ is harder to compute, however:
$$ \frac{\partial\mathcal{L}}{\partial \alpha} = \sum_{d}{\frac{\partial }{\partial \alpha} \E_{q}\left[\log p(z_d~|\alpha)\right]} = \sum_{d}{\E_{q}\left[\frac{\partial }{\partial \alpha} \log p(z_d~|\alpha)\right]} . $$
Since the expectation cannot be analytically computed, we use our samples $z_{d}^{(s')}$ for $s' = 1, \dots, S'$ and $d \in \mathcal{M}$ to compute a stochastic gradient. For this,
we first note that for fixed $z_d$,
\begin{align*}
\log p(z_d~|\alpha) & = \sum_{r}{\left[\log\Gamma(O_{dr} + \alpha) - \log\Gamma(\alpha)\right]} \\ 
                              &~~~~~~~~~~ + \log\Gamma(R\alpha) - \log\Gamma(O_{d} + R\alpha) , 
\end{align*}
whence
\begin{align*}
\frac{\partial }{\partial \alpha} \log p(z_d~|\alpha) & = \sum_{r}{\left[\Psi(O_{dr} + \alpha) - \Psi(\alpha)\right]}  \\
                                                                           &~~~~~~~~~ + R \cdot \left[\Psi(R\alpha) - \Psi(O_{d} + R\alpha)\right] \\
                                                                           &~~~~~~~~~~~~~~~~~~~ = \colon \hat{g}_{\alpha}(z_d), 
\end{align*}
where $O_{dr}$ denotes the number of sentences in $d$ assigned to relation $r$ and $O_d = \sum_r O_{dr}$. We thus obtain a stochastic gradient
$$ \hat{g}_{\alpha} = \frac{D}{S} \cdot \left[\sum_{d \in \mathcal{M}}{\frac{1}{S'}\sum_{s' = 1}^{S'}{\hat{g}_{\alpha}\left(z_{d}^{(s')}\right)}}\right] . $$